\documentclass[letterpaper]{article} %
\usepackage{aaai24}  %
\usepackage{times}  %
\usepackage{helvet}  %
\usepackage{courier}  %
\usepackage[hyphens]{url}  %
\usepackage{graphicx} %
\usepackage{natbib}  %
\usepackage{caption} %
\usepackage{algorithm}
\usepackage{algorithmic}

\usepackage{newfloat}
\usepackage{listings}
\DeclareCaptionStyle{ruled}{labelfont=normalfont,labelsep=colon,strut=off} %
\floatstyle{ruled}
\newfloat{listing}{tb}{lst}{}
\floatname{listing}{Listing}
\title{\model: A Neural Framework for Attributed Graph Clustering via Modularity Maximization}
\author {
    Aritra Bhowmick\textsuperscript{\rm 1},
    Mert Kosan\textsuperscript{\rm 2}\thanks{Work done prior to joining Visa Research.},
    Zexi Huang\textsuperscript{\rm 2},
    Ambuj Singh\textsuperscript{\rm 2},
    Sourav Medya\textsuperscript{\rm 3}
}
\affiliations {
    \textsuperscript{\rm 1}New York University\\
    \textsuperscript{\rm 2}University of California, Santa Barbara\\
    \textsuperscript{\rm 3}University of Illinois, Chicago\\
    aritra.bhowmick@nyu.edu, mertkosan@gmail.com, zexihuang.phd@gmail.com, ambuj@ucsb.edu, medya@uic.edu
}

\usepackage{amsmath}
\usepackage{comment}
\usepackage{amsfonts}
\usepackage{booktabs}

\newcommand{\model}{\textsc{DGCluster}{}}

\begin{document}

\maketitle

\begin{abstract}

Graph clustering is a fundamental and challenging task in the field of graph mining where the objective is to group the nodes into clusters taking into consideration the topology of the graph. It has several applications in diverse domains spanning social network analysis, recommender systems, computer vision, and bioinformatics.  In this work, we propose a novel method, {\model}, which primarily optimizes the modularity objective using graph neural networks and scales linearly with the graph size. Our method does not require the number of clusters to be specified as a part of the input and can also leverage the availability of auxiliary node level information. We extensively test {\model} on several real-world datasets of varying sizes, across multiple popular cluster quality metrics. Our approach consistently outperforms the state-of-the-art methods, demonstrating significant performance gains in almost all settings.

\end{abstract}

\section{Introduction and Related Work}

Graph clustering is a fundamental problem in network analysis and plays an important role in uncovering structures and relationships between the nodes or entities in a graph. It has numerous applications in several domains such as community detection in social networks \cite{newman2006finding}, identifying functional modules in biological systems \cite{wang2010fast}, image segmentation in computer vision~\cite{felzenszwalb2004efficient}, and recommender systems \cite{moradi2015effective}. The primary goal of graph clustering is to group nodes with similar characteristics or functions while maintaining a clear distinction between different clusters.

 \paragraph{Node attributes. }While traditional graph clustering methods primarily rely on graph topology such as modularity maximization \cite{newman2006modularity,newman2003mixing}, recent research \cite{wang2017mgae} has recognized the importance of incorporating node attributes in the clustering process, offering a more comprehensive approach for grouping nodes. Node attributes provide additional information associated with each node and provide contextual insights that can improve the accuracy in the clustering process. %

 \paragraph{GNN-based clustering. }Recently, there have been several attempts that use the power of deep learning in the form of Graph neural networks (GNNs) \cite{kipf2016semi, hamilton2017inductive, velickovic2017graph} for graph clustering. GNNs provide a powerful tool in graph-based machine learning that is successful in many diverse prediction tasks \cite{zhang2018link,ying2018hierarchical,ying2018graph},
by incorporating the graph topology node features or attributes. For GNN-based graph clustering, MGAE \cite{wang2017mgae} marginalizes the corrupted node features to learn representations via a graph encoder and applies spectral clustering. Another graph autoencoder based approach has been proposed in \cite{park2019symmetric}. To improve the efficiency of clustering, contrastive learning methods have been used recently as well \cite{liu2023simple,kulatilleke2022scgc,xia2021self}. For more neural methods on deep graph clustering, we refer the readers to this recent survey \cite{yue2022survey}. %

\paragraph{Neural modularity maximization.} One of the initial methods to optimize modularity via deep learning for graph clustering is proposed by \citet{yang2016modularity}. They design a nonlinear reconstruction method based on graph autoencoders, which also incorporate constraints among node pairs. \citet{wu2020deep} propose a method that obtains a spatial proximity matrix by using the adjacency matrix and the opinion leaders in the social network. The spatial eigenvectors of the proximity matrix are applied subsequently to optimize modularity. \citet{mandaglio2018consensus} is another study that incorporates the modularity metric for community detection and graph clustering. \citet{choong2018learning, bhatia2018dfuzzy} try to find communities without predefined community structure. \citet{choong2018learning} propose a generative model for community detection using a variational autoencoder. \citet{bhatia2018dfuzzy} firstly analyze the possible number of communities in the graph, then fine-tune it using modularity. Later on, \citet{sun2021graph} use a graph neural network that optimizes modularity and attributes similarity objectives. Another related work in this domain is the method DMoN by \citet{muller2023graph}. This method designs an architecture to encode cluster assignments and then formulate a modularity-based objective function for optimizing these assignments.

\paragraph{Our contributions. } A vast majority of these methods have the limitation that they require the number of clusters to be given as an input or they do not take full advantage of the associated node attributes along with the additional node level information like partial availability of labels or samples of known pairwise memberships. To overcome these challenges, we propose a framework named Deep Graph Cluster ({\model}) that eliminates the need for a predefined number of clusters and harnesses graph representation learning methods that can leverage node attributes along with other available auxiliary information. Our major contributions are as follows.\\

\begin{itemize}
    \item \textbf{\model:} We develop a novel framework that uses pairwise (soft) memberships between nodes to solve the graph clustering problem via modularity maximization. The complexity of our framework scales linearly with the size of the graph. 

    \item \textbf{Handling Unknown Number of Clusters and Auxiliary Information:} Our proposed methodology can generalize well to cases when the number of clusters is not known beforehand. Our designed loss function is also flexible towards accommodating the additional local or node-level information. These are the major strengths of our approach. 
 
    \item \textbf{Extensive Empirical Evaluation:} We conduct extensive experiments on seven real-world datasets of different sizes on four different objectives that quantify the quality of clusters. Our method shows significant performance gain against state-of-the-art methods in most of the settings.
    
\end{itemize}

\section{Problem Definition}
\label{sec:prob}

We consider an undirected and unweighted graph $G = (V, E)$,
where $V=\{v_1,v_2,v_3,...,v_n\}$ is the set of $n$ vertices/nodes, and
$E = \{e_{ij}=(v_i,v_j)\}$ is the set of $m$ edges. The adjacency matrix of $G$ can be represented as a non-negative symmetric
matrix $A = [A_{ij} ] \in R_+^{n \times n}$ where $A_{ij} = 1$ if there is an edge between vertices $i$ and $j$, and $A_{ij} = 0$. %
The degree of vertex $i$ is defined as $d_i = \sum_j a_{ij}$. 
In addition, we have features (or attributes) associated with each node in the graph, $X^0 \in \mathbb{R}^{n\times r}$ where $r$ is the size of the feature vector on nodes.  %

\paragraph{Graph clustering. }Our objective is to have a disjoint clustering of the nodes in the graph. More specifically, the problem of \textit{graph clustering} is to partition the set of nodes into $k$ clusters or groups $\{V_i \}^k_{i=1}$ ($V_i \cap V_j =\emptyset$  for $i \ne j$), %
such that the nodes within a cluster are more densely connected than nodes belonging to different clusters.  Furthermore, in this work, we aim to incorporate node attributes in addition to the graph topology for graph clustering. %

While there exist several metrics to measure the quality of clustering such as conductance \cite{yang2012defining} and normalized cut-ratio \cite{shi2000normalized}, modularity remains the most popular and widely used metric for graph clustering in the literature \cite{fortunato2016community}. 

\paragraph{Modularity. }The approach of graph clustering based on maximizing the modularity of the graph has been introduced by Newman \cite{newman2006modularity}. As a graph topology-based measure, modularity \cite{newman2006modularity} quantifies the difference between the fraction of the edges that fall within the clusters and the expected fraction assuming the edges have been distributed randomly. Formally, modularity ($Q$) is defined as follows:
\begin{equation}
     Q=\frac{1}{2m}\sum_{ij}(A_{ij}-\frac{d_id_j}{2m})\delta(c_i,c_j)
\end{equation}
where $\delta(c_i ,c_j)$ is the Kronecker delta, i.e., $\delta(c_i,c_j)=1$ if $c_i=c_j$ and $0$ otherwise,
and $c_i$ is the community to which node $i$ is
assigned. 

The value of modularity for unweighted and undirected graphs lies in the range $[-1/2,1]$. %
The $Q$ value close to $0$ implies
that the fraction of edges inside communities is no better than a random distribution, and higher values usually correspond to a stronger cluster structure.
The modularity $Q$ can also be expressed in the matrix form as follows:
\begin{equation}
\label{eq:q_matrix}
    Q= \frac{1}{2m} \sum_{ij} B \odot M = \frac{1}{2m} Tr(B M^T)=\frac{1}{2m} Tr(B M)    
\end{equation}
where $M$ is a $n\times n$ symmetric matrix with $M_{ij} = \delta (c_i, c_j)$ and $B_{ij}= (A_{ij}-\frac{d_id_j}{2m})$ is called the modularity matrix.\\

\textbf{Modularity maximization. } %
Since a larger $Q$ implies a prominent cluster structure, optimizing the modularity is a popular way of finding good clusters. While modularity optimization is known to be NP-Hard \cite{brandes2006maximizing}, there exist techniques such as spectral relaxation and greedy algorithms, which permit efficient solutions \cite{newman2006finding,blondel2008fast}.

\textbf{Our goal.} We achieve graph clustering via modularity maximization. The definition of modularity brings the idea of computing pairwise memberships allowing a natural interpretation without knowing the number of clusters. We aim to take advantage of that and the power of graph representation learning techniques that can exploit both structural and non-structural information from graphs. In the experiments, we show the efficacy of our method on four different objectives that quantify the quality of the clusters: modularity \cite{newman2006modularity}, conductance \cite{yang2012defining}, Normalized mutual information (NMI), and F1 score. %

\section{Method: \model}

We present \model, a fully differentiable method which performs deep graph clustering based on the graph structure and node attributes, without the need to pre-define the number of communities.
The key rationale of our method is to parameterize a relevant clustering objective (e.g., Modularity) with similarity between nodes computed based on graph neural network (GNN)-based embeddings. Our method {\model} consists of four steps:
\begin{itemize}
    \item \textbf{Node embeddings. }As the first step, {\model} obtains the node embeddings as the output of the GNN followed by some transformations that helps to perform efficient clustering.
    \item 
    \textbf{Modularity via similarity. }We evaluate the similarity between all node pairs from the embeddings and treat them as soft community pairwise memberships.%
    \item \textbf{Objectives. }Subsequently, it builds the cluster detection objectives based on the soft memberships and train the GNN parameters in a differentiable manner.%
    \item \textbf{Final clustering. } Finally, it computes the community memberships by clustering the GNN node embeddings.%
\end{itemize}

We introduce each step of our method {\model} in the following subsections. 

{\subsection{Node embeddings using GNN}\label{subsec::node_embedding}} 

We begin with a brief introduction of graph neural networks (GNNs). GNNs are powerful graph machine learning paradigms that combine the graph structure and node attribute information into node embeddings for different downstream tasks. %
A key design element of GNNs is message passing where the nodes iteratively update their representations (embeddings) by aggregating information from their neighbors. In the literature, several different GNN architectures have been proposed~\cite{scarselli2008graph, kipf2016semi, hamilton2017inductive, velickovic2017graph} which implement different schemes of message passing. A comprehensive discussion on the methods and applications of GNNs are described here \cite{zhou2020graph}.

In this paper, we leverage the widely used Graph Convolutional Network (GCN)~\cite{kipf2016semi} to produce node embeddings, noting that our model can be equipped with other GNNs. With the initial node features as $X^{(0)}$, the layer-wise message passing rule for layer $l$ ($l=0,\cdots, L-1$) is as follows: 

\begin{equation}
    X^{(l+1)}=\sigma ( \Tilde{A} X^{(l)}W^{(l)})
\end{equation}
where $\Tilde{A}=D^{-\frac{1}{2}}AD^{\frac{1}{2}}$ is the normalized adjacency matrix, $D$ is the diagonal node degree matrix, $X^l$ is the embedding output of the $l$-th layer, $W^l$ is the learnable weight matrix of the $l$th layer, and $\sigma$ is the activation function which introduces the non-linearity in the feature aggregation scheme. We do not use any self loop creation in the adjacency matrix and we choose the SELU \cite{klambauer2017self} as the activation function for better convergence. The SELU activation is given as:
\begin{equation}
    SELU(x)=
    \begin{cases}
        \beta x,& \text{if } x \geq 0\\
        \beta \alpha(e^x-1),& \text{otherwise}
    \end{cases}
\end{equation} 
where $\beta \approx 1.05$ and $\alpha \approx 1.67$.\\

\textbf{Transformation of the embeddings. }Let $X=X^{L}=[X_1, \cdots, X_n]^T$ be the output embeddings of the last layer readout from the GNN. We introduce the following problem-specific transformations on the embeddings (where $Z^{\circ 2}$ denotes the element-wise square operation): %
\begin{equation}
 \begin{aligned}
X_i&\leftarrow\frac{X_i}{\sum_j X_{ij}},\\
X_i&\leftarrow\tanh(X_i),\\
 X_i &\leftarrow X_i^{\circ 2},\\
X_i&\leftarrow\frac{X_i}{\|X_i\|_2}    
\end{aligned}   
\end{equation}

Specifically, the first normalization is used so as to prevent vanishing gradients because of the $\tanh$ activation function for large values. Next, after the activation, doing the element-wise square ensures the output is constrained within the positive coordinate space. The final $L_2$ normalization reduces the cosine similarity computation of node pairs (introduced in next subsection) to corresponding dot products. 
Thus, the final embeddings %
lie on the surface of the unit sphere constrained in the positive space. More detailed intuitions behind these will be explained in details in the following sections.\\

\subsection{Modularity via embedding similarity}
\label{subsec::similarity_computation}

After obtaining the transformed GNN embeddings $X$, we demonstrate how to compute the modularity $Q$ based on them via pairwise node similarities.%

To achieve this, we focus on $M_{}$ which is defined in Eq.~\ref{eq:q_matrix} as a binary matrix that encodes the pairwise memberships (via $\delta$) of the nodes in a cluster. This pairwise relationship is transitive, i.e., $\delta(c_u,c_v)=1$ and $\delta(c_v,c_w)=1$ implies $\delta(c_u,c_w)=1$.  %
However, as stated before, the problem of finding the optimal $M_{}$ which maximizes $Q$ is NP-Hard. The main idea is to replace $M_{}$ with a soft pairwise community membership matrix. We choose to replace $M_{}$ with a similarity matrix which is defined based on node embeddings, where the similarity ($f_{sim}(X_u, X_v)\in [0,1]$) can be viewed as soft membership between the nodes $\{u,v\}$.  Higher values of $f_{sim}(X_u, X_v)$ corresponds to higher similarity or stronger relationship between the nodes.  %
Although there can be many choices for the similarity function, we choose $f_{sim}$ as the cosine similarity, $f_{sim}(X_u, X_v) = \cos(X_u, X_v)$. %

Here, we also emphasize our rationale for the transformations of the embeddings earlier. Specifically, the original $M_{}$ only takes binary values (i.e., 0 or 1). Our embedding transformation allows $\cos(X_u, X_v)\in [0,1]$ by limiting them in the positive coordinate space, and enables its computation via dot products $\cos(X_u, X_v) = X_u^TX_v$, which in turn leads to an efficient computation as discussed later.

\subsection{Objective function}
\label{subsec::objective_function}

We introduce a novel joint objective function that performs clustering based on both the community structure quality measure (i.e., modularity) and local auxiliary information.

\subsubsection{Modularity optimization.}
We first define our primary objective function, i.e., modularity optimization by approximating the pairwise community membership computed with embedding similarity:%

\begin{equation}
    L_1=-\Tilde{Q} = -\frac{1}{2m}Tr(BXX^T) 
\end{equation}
where $B$ is the modularity matrix, $m$ is the number of edges, and $XX^T$ is the similarity matrix obtained from the transformed node embeddings.

\subsubsection{Auxiliary information loss.}
We now discuss how our algorithm can be made more flexible by accounting for additional local information. %

Specifically, let $S \subseteq V$ be the subset of nodes whose local information is available, and $H \in \mathbb{R}^{|S| \times |S|}$ be the pairwise information matrix. We consider different types of additional information. In the semi-supervised setting, partial node labels (e.g. cluster labels, class labels) are available, and we can construct $H$ based on the pairwise membership: 
\begin{equation}
    H_{ij}=
    \begin{cases}
        1,& \text{if } c_i=c_j\\
        0,& \text{otherwise}
    \end{cases}
\end{equation}
where $c_i$ is the label of the node $i$. Alternatively, by collecting all available labels in one-hot matrix form $C \in \mathbb{R}^{|S| \times p}$, we can rewrite $H$ as: 
\begin{equation}
    H = CC^T
\end{equation}

When those ground-truth information is not available, we can also leverage traditional structure-based graph partitioning heuristics, such as Louvain \cite{blondel2008fast}, and treat their generated clusters as the node labels. 
In general, any pairwise node information similarity which can be approximated as $\langle C_i, C_j\rangle \in [0,1]$, can be effectively used.

With the local information matrix $H$, the secondary objective function which minimizes the difference between $H$ and the embedding similarity matrix is given as :
\begin{equation}
    L_2 = \frac{1}{|S|^2}\|H-X_SX_S^T\|_F^2
\end{equation}
where $X_S$ is the submatrix of node embeddings with only nodes in $S$. 

The final objective function is a weighted combination of the two objectives:
\begin{equation}
    L = L_1 + \lambda L_2
\end{equation}
where $\lambda$ is a hyperparameter. The GNN parameters are trained in an end-to-end based on the total loss $L$ with the stochastic gradient descent algorithm. 

When instead of individual node labels, samples of pairwise node memberships are available, $L_2$ can be written as follows:
\begin{equation}
    L_2=\frac{1}{|S|}\sum_{p_{ij} \in S}(1-\langle X_i, X_j \rangle)^2    
\end{equation}
where $p_{ij}$ are the node pairs which belong to the same community and $S$ is a set of such pairs.

\subsection{Clustering node embeddings}
\label{subsec::final_clustering}
In this section, we illustrate how to obtain the hard cluster partitions based on the soft pairwise memberships obtained in the previous section. 

One way is to directly apply clustering algorithms based on the pairwise node similarity matrix, such as affinity propagation  \cite{frey2007clustering}. However, computing the full similarity matrix from the embeddings is computationally prohibitive. Instead, we take advantage of the following observation. Since our embeddings are $L_2$ normalized (i.e., $\|X_u\|_2=1$), the cosine similarity is directly related to the Euclidean distance in the embedding space:
\begin{equation}
    \small
    \begin{aligned}
    \|X_u-X_v\|_2^2&=\|X_u\|_2^2+\|X_v\|_2^2-2\|X_u\|_2\|X_v\|_2\cos(X_u, X_v)\\
    &=2(1-\cos(X_u, X_v))        
    \end{aligned}
\end{equation}
This allows us to apply clustering algorithms based on the euclidean distance in the embedding space without computing the full pairwise similarity matrix.

Specifically, we apply the Balanced Iterative Reducing and Clustering using Hierarchies (BIRCH) \cite{zhang1996birch}. BIRCH is a scalable, memory-efficient, online clustering algorithm that can cluster large datasets by first generating a small and compact summary of the dataset that retains as much information as possible. Unlike other popular choices such as k-means \cite{hartigan1979algorithm}, BIRCH does not require the number of clusters beforehand.

\subsection{Complexity analysis}
\label{subsec::complexity}
In this section, we analyze the complexity of our proposed model. The forward pass requires us to compute two objective functions. Specifically, modularity optimization loss $L_1$ can be evaluated with

\begin{equation}
\small
    \begin{aligned}
          L_1 &= -\frac{1}{2m} Tr(B XX^T) =-\frac{1}{2m} Tr(X^T B X)\\
          &=-\frac{1}{2m}( Tr(X^TAX) - \frac{1}{2m}Tr(X^Tdd^TX) )  
    \end{aligned}  
\end{equation}
We can see that $L_1$ can be computed with sparse matrix multiplications between $X$ and $A$ and matrix vector multiplications between $X$ and $d$. These multiplications lead to an overall computation cost of $O(k^2n)$, where $k$ is the dimension of the embeddings $X$. %

For the auxiliary information loss, we have

\begin{equation}
\small
    \begin{aligned}
        &\|H-X_SX_S^T\|_F^2=\sum_{ij}(H_{ij}-(XX^T)_{ij})^2\\
        &= \sum_{ij}H_{ij}^2+\sum_{ij}(XX^T)_{ij}^2
        - 2 \sum_{ij}H_{ij}(XX^T)_{ij}  \\
        &= \sum_{ij}(CC^T)_{ij}^2+\sum_{ij}(XX^T)_{ij}^2
        - 2 \sum_{ij}(CC^T)_{ij}(XX^T)_{ij}  \\
        &=Tr(C^TCC^TC)-Tr(X_S^TX_SX_S^TX_S)- 2 Tr(X_S^TCC^TX_S )\\
    \end{aligned}
\end{equation}

Computing $C^TC$, $X_S^TX_S$, and $X_S^TC$ requires $O(p^2n)$, $O(k^2n)$, and $O(knp)$ respectively via matrix multiplications, assuming $|S|=n$, since $|S|$ can be atmost $n$. Here, $p$ is the dimension of the auxiliary information (note, $C \in \mathbb{R}^{n \times p}$ ). Thus, the overall complexity of our model is $O(k^2n+p^2n)$, where $k,p \ll n$. This shows that our model scales linearly with the size of the graph. %

\section{Experimental Results}
\label{sec::experiments}

In this section, we present the performance comparison of {\model} with other clustering and neural community detection methods on \textit{$7$ well-known real-world datasets}. We evaluate the algorithms using various metrics, including \textit{graph conductance, modularity, NMI (Normalized Mutual Information), and F1 score} where higher scores are desired for the last three. Furthermore, we demonstrate the robustness of our method against different GNN architectures and assess the impact of adding auxiliary information. We also introduce an additional regularization objective and show its effect on our method. Lastly, we illustrate the number of communities our method identifies for each dataset. Our code and implementation of {\model} is available at https://github.com/pyrobits/DGCluster

\subsubsection{Datasets.}

We use seven publicly available real-world datasets in our experiments. Table \ref{tab::dataset} presents a summary of the dataset statistics. Our experiments include datasets from well-known citation networks, namely \textsc{Cora}, \textsc{Citeseer}, and \textsc{PubMed} \cite{sen2008collective}. In these networks, nodes correspond to individual papers, edges represent citations between papers, and features are extracted using a bag-of-words approach applied to paper abstracts. The topic of each paper is captured through node labels. Besides citation networks, we use two co-purchase networks, \textsc{Amazon PC} and \textsc{Amazon Photo} \cite{shchur2018pitfalls}. These networks include products as nodes, with edges denoting co-purchase relationships. Features are extracted from product reviews, while node labels show product categories. Our last two datasets, \textsc{Coauthor CS} and \textsc{Coauthor PHY} \cite{shchur2018pitfalls, shchur2019overlapping}, are co-authorship networks for computer science and physics, respectively. Within these networks, nodes correspond to authors, and edges indicate co-authorship between them. Node features are keywords extracted from authors' publications and node labels are their fields of study.

\begin{table}[htbp]
\small
\centering
\begin{tabular}{ccccc}
\toprule
 & $|V|$ & $|E|$ & $|X|$ & $|Y|$ \\ \midrule
\textsc{Cora}             & 2708         & 5278         & 1433         & 7            \\
\textsc{Citeseer}         & 3327         & 4552         & 3703         & 6            \\
\textsc{Pubmed}           & 19717        & 44324        & 500          & 3            \\
\textsc{Amazon PC}        & 13752        & 245861       & 767          & 10           \\
\textsc{Amazon Photo}     & 7650         & 119081        & 745          & 8            \\
\textsc{Coauthor CS}      & 18333        & 81894        & 6805         & 15           \\
\textsc{Coauthor PHY}    & 34493        & 247962       & 8415         & 5            \\
\bottomrule
\end{tabular}
\caption{Statistics of the datasets. $|V|$, $|E|$, $|X|$, and $|Y|$ denote the number of nodes, edges, features, and node labels.}
\label{tab::dataset}
\end{table}

\subsubsection{Baselines.}
Our baseline methods consist of a range of classical clustering algorithms, such as k-means, as well as state-of-the-art graph community detection algorithms. For consistent comparison, we adopted the same baseline setting as in the recent \textbf{DMoN} \citep{muller2023graph}: \textbf{k-means} based on features, \textbf{SBM} by \citet{peixoto2014efficient}, \textbf{k-means} based on DeepWalk by \citet{perozzi2014deepwalk}, \textbf{k-means}(DGI) by \citet{velivckovic2018deep}, \textbf{AGC} by \citet{zhang2019attributed}, \textbf{DAEGC} by \citet{wang2019attributed}, \textbf{SDCN} by \citet{bo2020structural}, \textbf{NOCD} by \citet{shchur2019overlapping}, \textbf{DiffPool} by \citet{ying2018graph}, \textbf{MinCutPool} by \citet{bianchi2020spectral}, and \textbf{Ortho} by \citet{bianchi2020spectral}.  %

\subsubsection{Performance Measures.}
We use our primary objective, modularity, as our main evaluation metric. Additionally, we evaluate the performance using other important metrics, namely NMI, conductance, and F1 score. In cases where a dataset lacks ground truth cluster labels, we assign node labels as their respective cluster labels. We multiply the metrics by 100 for better readability.

\begin{itemize}
    \item Modularity ($Q$) \cite{newman2006modularity}: This serves as the primary objective, aiming to quantify the strength of the community. It achieves this by contrasting the proportion of edges within the community with a corresponding fraction generated from random connections between nodes. Higher values mean better community partitions.
    \item Conductance ($C$) \cite{yang2012defining}: The graph's conductance measures the portion of total edges that goes outside the community. Lower values indicate better community partitions.
    \item Normalized mutual information (NMI): We use the NMI as a label-based metric between the cluster assignments and true labels of the nodes. Higher values mean better community partitions.
    \item F1 score: We calculate the pairwise F1 score based on found pairwise node memberships and their corresponding clusters. Since it is $O(N^2)$, we sample 1000 nodes to calculate this. Higher values indicate better community partitions.
\end{itemize}

\paragraph{Other Settings.}

In our experiments, we employ a GCN having two hidden layers of 256 and 128 nodes and an output dimension of 64, across all datasets. We showcase outcomes for various $\lambda$ values. This enables a comparative analysis among the graph structure-based metrics and graph attribute-based metrics. We use Adam optimizer with a learning rate set to 0.001, and we set the number of epochs to 300. Our tables and figures report average scores over 10 different runs, using different seeds, for our method.
For baselines, we report the results from \citet{muller2023graph}.

\begin{table*}[htbp]
\centering
\resizebox{0.97\textwidth}{!}{
\begin{tabular}{ccccccccccccccc}
\toprule
 & \multicolumn{2}{c}{\textsc{Cora}} & \multicolumn{2}{c}{\textsc{Citeseer}} & \multicolumn{2}{c}{\textsc{Pubmed}} & \multicolumn{2}{c}{\textsc{Amazon PC}} & \multicolumn{2}{c}{\textsc{Amazon Photo}} & \multicolumn{2}{c}{\textsc{Coauthor CS}} & \multicolumn{2}{c}{\textsc{Coauthor PHY}} \\ \midrule
method & $C \downarrow$ & $Q \uparrow$ & $C \downarrow$ & $Q \uparrow$ & $C \downarrow$ & $Q \uparrow$ &  $C \downarrow$ & $Q \uparrow$ & $C \downarrow$ & $Q \uparrow$ & $C \downarrow$ & $Q \uparrow$ & $C \downarrow$ & $Q \uparrow$ \\ \midrule
k-m(feat) & $61.7$ & $19.8$ & $60.5$ & $30.3$ & $55.8$ & $33.4$ & $84.5$ & $5.4$ & $79.6$ & $10.5$ & $49.1$ & $23.1$ & $57.0$ & $19.4$ \\
SBM & $15.4$ & $77.3$ & $14.2$ & $78.1$ & $39.0$ & $53.5$ & $31.0$ & $\underline{60.8}$ & $18.6$ & $\textbf{72.2}$ & 20.3 & $\underline{72.7}$ & $25.9$ & $\underline{66.9}$ \\
k-m(DW) & $62.1$ & $30.7$ & $68.1$ & $24.3$ & $\textbf{16.6}$ & $\textbf{75.3}$ & $67.6$ & $11.8$ & $60.6$ & $22.9$ & $33.1$ & $59.4$ & $44.7$ & $47.0$ \\
AGC & $48.9$ & $43.2$ & $41.9$ & $50.0$ & $44.9$ & $46.8$ & $43.2$ & $42.8$ & $35.3$ & $33.8$ & $41.5$ & $40.1$ & N/A & N/A \\
SDCN & $37.5$ & $50.8$ & $20.0$ & $62.3$ & $22.4$ & $50.3$ & $25.1$ & $45.6$ & $19.7$ & $53.3$ & $33.0$ & $55.7$ & $32.1$ & $52.8$ \\
DAEGC & $56.8$ & $33.5$ & $47.6$ & $36.4$ & $53.6$ & $37.5$ & $39.0$ & $43.3$ & $19.3$ & $58.0$ & $39.4$ & $49.1$ & N/A & N/A \\
k-m(DGI) & $28.0$ & $64.0$ & $17.5$ & $73.7$ & $82.9$ & $9.6$ & $61.9$ & $22.8$ & $51.5$ & $35.1$ & $35.1$ & $57.8$ & $38.6$ & $51.2$ \\
NOCD & $14.7$ & $\underline{78.3}$ & $6.8$ & $\underline{84.4}$ & $21.7$ & $69.6$ & $26.4$ & $59.0$ & $13.7$ & $70.1$ & $20.9$ & $72.2$ & $25.7$ & $65.5$ \\
DiffPool & $26.1$ & $66.3$ & $26.0$ & $63.4$ & $32.9$ & $56.8$ & $35.6$ & $30.4$ & $26.5$ & $46.8$ & $33.6$ & $59.3$ & N/A & N/A \\
MinCutPool & $23.3$ & $70.3$ & $14.1$ & $78.9$ & $29.6$ & $63.1$ & N/C & N/C & N/C & N/C & $22.7$ & $70.5$ & $27.8$ & $64.3$ \\
Ortho & $28.0$ & $65.6$ & $18.4$ & $74.5$ & $57.8$ & $32.9$ & N/C & N/C & N/C & N/C & $27.8$ & $65.7$ & $33.0$ & $59.5$ \\
DMoN & \underline{$12.2$} & $76.5$ & \underline{$5.1$} & $79.3$ & \underline{$17.7$} & $65.4$ & \underline{$18.0$} & $59.0$ & \underline{$12.7$} & $70.1$ & \underline{$17.5$} & $72.4$ & $\textbf{18.8}$ & $65.8$ \\

\midrule

\model ($\lambda=0.0$) & $\textbf{8.4}$ & $\textbf{80.7}$ & $5.5$ & $87.2$ & $20.6$ & $72.5$ & $17.4$ & $61.3$ & $7.5$ & $70.9$ & $\textbf{14.6}$ & $\textbf{74.3}$ & $\underline{20.6}$ & $\textbf{67.5}$ \\
\model ($\lambda=0.2$) & $9.7$ & $80.8$ & $\textbf{4.1}$ & $\textbf{87.4}$ & $20.4$ & $\underline{72.8}$ & $\textbf{17.7}$ & $\textbf{61.5}$ & $\textbf{8.6}$ & $\underline{71.6}$ & $15.3$ & $74.2$ & $22.3$ & $67.3$ \\

\model ($\lambda=0.8$) & $14.5$ & $78.6$ & $6.5$ & $86.3$ & $24.6$ & $71.2$ & $27.3$ & $60.3$ & $12.4$ & $71.6$ & $18.1$ & $73.3$ & $21.3$ & $66.0$ \\

\bottomrule
\end{tabular}
}
\caption{Performance across datasets evaluated using graph conductance $C$ and graph modularity $Q$, with three $\lambda$ settings (0, 0.2, 0.8) from our method. The best and second-best methods are highlighted in \textbf{bold} and \underline{underlined} for each dataset-metric pair. We fixed the optimal $\lambda$ value per dataset in our method (i.e., 0.0 or 0.2) during comparison, and our method demonstrates the best or comparable performance to the baselines. Unavailable results and non-convergence are labeled as N/A and N/C.} %
\label{tab:graph_based_eval}
\end{table*}

\begin{table*}[h!]
\centering
\resizebox{0.97\textwidth}{!}{
\begin{tabular}{ccccccccccccccc}
\toprule
 & \multicolumn{2}{c}{\textsc{Cora}} & \multicolumn{2}{c}{\textsc{Citeseer}} & \multicolumn{2}{c}{\textsc{Pubmed}} & \multicolumn{2}{c}{\textsc{Amazon PC}} & \multicolumn{2}{c}{\textsc{Amazon Photo}} & \multicolumn{2}{c}{\textsc{Coauthor CS}} & \multicolumn{2}{c}{\textsc{Coauthor PHY}} \\ \midrule
method & NMI $\uparrow$ & F1 $\uparrow$ & NMI $\uparrow$ & F1 $\uparrow$ & NMI $\uparrow$ & F1 $\uparrow$ & NMI $\uparrow$ & F1 $\uparrow$ & NMI $\uparrow$ & F1 $\uparrow$ & NMI $\uparrow$ & F1 $\uparrow$ & NMI $\uparrow$ & F1 $\uparrow$ \\
\midrule
k-m(feat) & 18.5 & 27.0 & 24.5 & 29.2 & 19.4 & 24.4 & 21.1 & 19.2 & 28.8 & 19.5 & 35.7 & 39.4 & 30.6 & \underline{42.9} \\
SBM & 36.2 & 30.2 & 15.3 & 19.1 & 16.4 & 16.7 & 48.4 & 34.6 & 59.3 & 47.4 & 58.0 & 47.7 & 45.4 & 30.4 \\
k-m(DW) & 24.3 & 24.8 & 27.6 & 24.8 & 22.9 & 17.2 & 38.2 & 22.7 & 49.4 & 33.8 & \underline{72.7} & \underline{61.2} & 43.5 & 24.3 \\
AGC & 34.1 & 28.9 & 25.5 & 27.5 & 18.2 & 18.4 & \underline{51.3} & 35.3 & 59.0 & 44.2 & 43.3 & 31.9 & N/A & N/A \\
SDCN & 27.9 & 29.9 & 31.4 & \underline{41.9} & 19.5 & 29.9 & 24.9 & 45.2 & 41.7 & 45.1 & 59.3 & 54.7 & 50.4 & 39.9 \\
DAEGC & 8.3 & 13.6 & 4.3 & 18.0 & 4.4 & 11.6 & 42.5 & 37.3 & 47.6 & 45.0 & 36.3 & 32.4 & N/A & N/A \\
k-m(DGI) & \underline{52.7} & 40.1 & \underline{40.4} & 39.4 & 22.0 & 26.4 & 22.6 & 15.0 & 33.4 & 23.6 & 64.6 & 51.9 & 51.0 & 30.6 \\
NOCD & 46.3 & 36.7 & 20.0 & 24.1 & 25.5 & 20.8 & 44.8 & 37.8 & 62.3 & 60.2 & 70.5 & 56.4 & 51.9 & 28.7 \\
DiffPool & 32.9 & 34.4 & 20.0 & 23.5 & 20.2 & 26.3 & 22.1 & 38.3 & 35.9 & 41.8 & 41.6 & 34.4 & N/A & N/A \\
MinCut & 35.8 & 25.0 & 25.9 & 20.1 & 25.4 & 15.8 & N/C & N/C & N/C & N/C & 64.6 & 47.8 & 48.3 & 24.9 \\
Ortho & 38.4 & 26.6 & 26.1 & 20.5 & 20.3 & 13.9 & N/C & N/C & N/C & N/C & 64.6 & 46.1 & 44.7 & 23.7 \\
DMoN & 48.8 & \underline{48.8} & 33.7 & \textbf{43.2} & \underline{29.8} & \underline{33.9} & 49.3 & \underline{45.4} & \underline{63.3} & \underline{61.0} & 69.1 & 59.8 & \underline{56.7} & 42.4 \\

\midrule
\model ($\lambda=0.0$) & 49.9 & 42.0 & 26.3 & 20.0 & 24.9 & 29.0 & 51.0 & 47.5 & 68.0 & 64.3 & 72.0 & 72.3 & 54.7 & 40.7 \\
\model ($\lambda=0.2$) & 53.0 & 43.5 & 30.3 & 22.2 & 27.6 & 30.1 & 53.8 & 49.5 & 73.0 & 70.7 & 76.1 & 77.3 & 59.0 & 41.9 \\
\model ($\lambda=0.8$) & \textbf{62.1} & \textbf{54.5} & \textbf{41.0} & 32.2 & \textbf{32.6} & \textbf{34.6} & \textbf{60.4} & \textbf{52.2} & \textbf{77.3} & \textbf{75.9} & \textbf{82.1} & \textbf{83.5} & \textbf{65.7} & \textbf{49.2} \\
\bottomrule
\end{tabular}
}
\caption{Performance across datasets evaluated using NMI and F1-score metrics, with three $\lambda$ settings (0, 0.2, 0.8) from our method. The best and second-best methods are highlighted in \textbf{bold} and \underline{underlined} for each dataset-metric pair. We fixed the optimal $\lambda$ value per dataset in our method (i.e., 0.8) during comparison, and our method demonstrates the best or comparable performance across most datasets. Unavailable results and non-convergence are labeled as N/A and N/C.}
\label{tab:label_based_eval}
\end{table*}

\begin{table*}[h!]
\centering
\resizebox{0.97\textwidth}{!}{
\begin{tabular}{ccccccccccccccccc}
\toprule
& \multicolumn{4}{c}{$C$} & \multicolumn{4}{c}{$Q$} & \multicolumn{4}{c}{NMI} & \multicolumn{4}{c}{F1} \\
& GCN & GAT & GIN & SAGE & GCN & GAT & GIN & SAGE & GCN & GAT & GIN & SAGE & GCN & GAT & GIN & SAGE \\
\midrule
\textsc{Cora} & 9.7 & 10.2 & 7.3 & 8.5 & 80.8 & 80.8 & 80.2 & 78.4 & 53.0 & 51.9 & 53.2 & 53.4 & 43.5 & 41.3 & 46.7 & \textbf{51.9} \\
\textsc{CiteSeer} & 4.1 & 4.1 & 4.5 & 4.8 & 87.4 & 87.4 & 87.3 & 84.8 & 30.3 & 29.4 & 29.5 & 32.5 & 22.2 & 21.9 & 21.9 & \textbf{34.0} \\
\textsc{PubMed} & 20.4 & 19.1 & 15.6 & \textbf{10.4} & 72.8 & 73.8 & 75.0 & 72.4 & 27.6 & 26.1 & 25.7 & 27.9 & 30.1 & 26.8 & 29.2 & \textbf{40.5} \\
\textsc{Amazon PC} & 17.7 & \underline{23.4} & 21.3 & \textbf{11.1} & 61.5 & 61.4 & 54.3 & \underline{55.1} & 53.8 & 52.9 & \underline{35.4} & \underline{45.0} & 49.5 & 45.3 & \underline{42.3} & 52.2 \\
\textsc{Amazon Photo} & 8.6 & 8.5 & 6.9 & 5.2 & 71.6 & 72.1 & 69.9 & \underline{63.8} & 73.0 & 72.1 & \underline{61.8} & \underline{60.6} & 70.7 & 69.4 & \underline{60.6} & \underline{58.2} \\
\textsc{Coauthor CS} & 15.3 & 15.7 & 14.7 & 13.6 & 74.2 & 74.0 & 73.8 & 73.0 & 76.1 & 75.7 & 71.8 & 72.4 & 77.3 & 79.0 & 72.6 & 74.4 \\
\textsc{Coauthor PHYSICS} & 22.3 & 20.3 & 18.5 & \textbf{14.5} & 67.3 & 67.1 & 66.7 & 66.0 & 59.0 & 59.2 & 59.7 & \textbf{64.0} & 41.9 & 43.4 & 48.7 & \textbf{55.9} \\
\midrule
\textsc{AVERAGE} & 14.0 & \underline{14.5} & 12.7 & \textbf{9.7} & 73.6 & \textbf{73.8} & 72.5 & \underline{70.5} & \textbf{53.3} & 52.5 & \underline{48.2} & 50.8 & 47.9 & 46.7 & \underline{46.0} & \textbf{52.4} \\

\bottomrule

\end{tabular}
}
\caption{Performance of our method with various GNN base models assessed across different metrics for all datasets, along with the calculation of average performance across datasets. Notably, the performance remains consistent across different variants of GNN models, demonstrating the absence of a clear winner. This shows the robustness of our model in adapting to diverse GNN architectures. For each dataset, we emphasize performance variations: a performance increase of more than 5 is highlighted in \textbf{bold}, while a decrease is marked with an \underline{underline}, both in comparison to GCN. Similarly, in the average row, we identify the best-performing GNN base model in \textbf{bold} and the worst-performing with an \underline{underline}, considering all metrics across datasets.}
\label{tab:base_model_effect}
\end{table*}

\subsection{Performance}

Table \ref{tab:graph_based_eval} and \ref{tab:label_based_eval} show that our method achieves superior or comparable results compared to the baselines across all evaluation metrics and datasets. Notably, our method stands out with significant improvements in unsupervised graph evaluation topology-based metrics such as graph conductance, and modularity (Table \ref{tab:graph_based_eval}). Our auxiliary objective also provides flexibility by tuning the hyperparameter $\lambda$ to optimize NMI and F1 score to have better results compared to the baselines (Table \ref{tab:label_based_eval}). While our method has a superior performance in co-purchase and co-authorship networks under all four metrics, it has a reasonably good performance in the citation networks. This shows the generalizability of our algorithm in different types of networks.

\subsection{Robustness on Different Base GNNs}

We conduct experiments to test the robustness of our method towards different GNNs including GAT \cite{velickovic2017graph}, GIN \cite{xu2018powerful}, and GraphSAGE \cite{hamilton2017inductive}. Table \ref{tab:base_model_effect} indicates that the selection of the GNN model does not significantly alter the performance of our method in terms of the evaluation measures. The average performance across datasets for different metrics and GNN bases reveals specific enhancements: GAT improves $Q$, GraphSAGE enhances $C$ and F1 scores on average. Conversely, GIN results into general performance degradation.

\begin{figure}[h]
    \centering
    \includegraphics[width=0.45\textwidth]{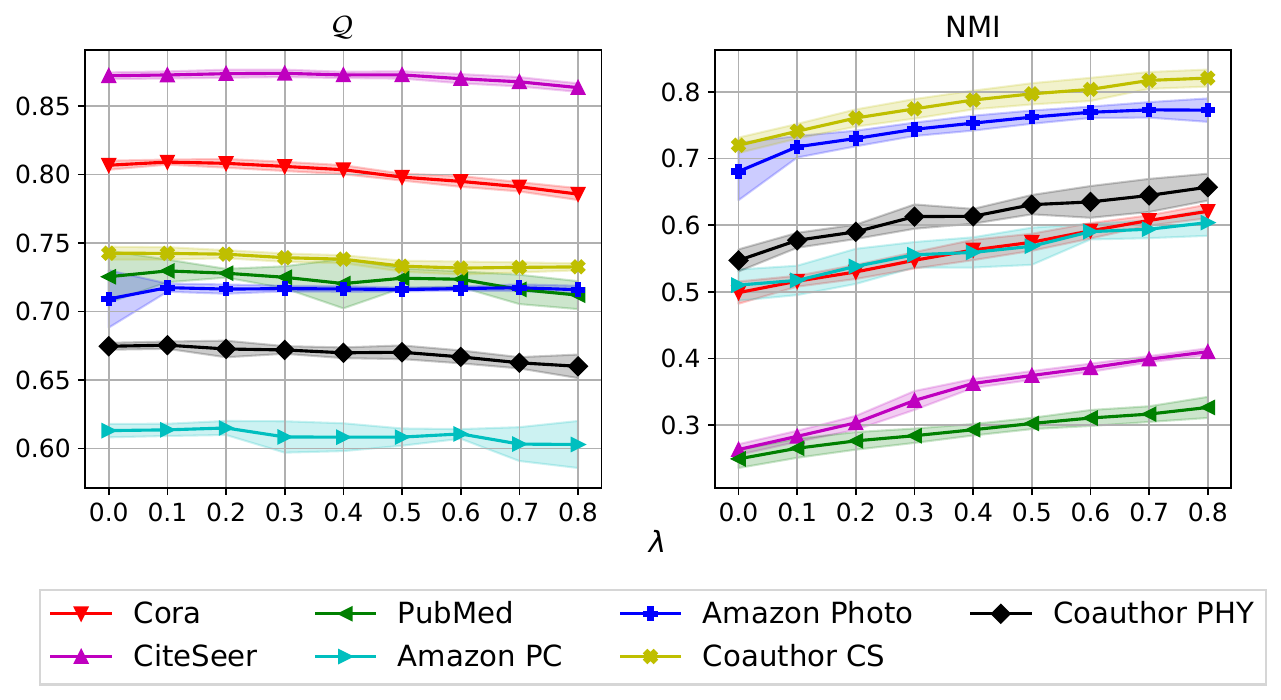}
    \caption{Effect of auxiliary information on different metrics and datasets with varying $\lambda$. As expected, $Q$ tends to decrease as $\lambda$ increases since $Q$ is dependent only on the graph structure whereas $\lambda$ adds weight to the label-based loss. The NMI increases as $\lambda$ increases since NMI is directly related to labels.}
    \label{fig:lambda_effect_on_modularity_nmi}
\end{figure}

\subsection{Auxiliary Information Effect}

To check the effect of our auxiliary information, we vary the $\lambda$ hyperparameter in our additional experiments. Figure \ref{fig:lambda_effect_on_modularity_nmi} illustrates the impact of the hyperparameter on different metrics and datasets. The main goal of the $\lambda$ parameter is to find the best underlying partition leveraging both graph structure and auxiliary information. We can see modularity $Q$ has a consistent value over increased $\lambda$, while NMI values are increasing with a good amount. This shows the power of our auxiliary objective which ensures flexibility to the user aiming to optimize their desired metric. Additionally, the standard deviation of our results is small showing the stability of our algorithm.

\begin{table}[htbp]
\centering
\resizebox{0.47\textwidth}{!}{
\begin{tabular}{ccccccc}
\toprule
 & \multicolumn{2}{c}{\textsc{Cora}} & \multicolumn{2}{c}{\textsc{Citeseer}} & \multicolumn{2}{c}{\textsc{Pubmed}} \\ \midrule
\model & $Q \uparrow$ & NMI $\uparrow$ & $Q \uparrow$ & NMI $\uparrow$ & $Q \uparrow$ & NMI $\uparrow$ \\
\midrule
$\alpha=0.0$ & 80.8 & 43.5 & 87.4 & 22.2 & 72.8 & 30.1 \\
$\alpha=0.5$ & 80.8 & 40.3 & 87.3 & 20.3 & 72.1 & 28.1 \\
$\alpha=1.0$ & 80.8 & 38.5 & 87.4 & 19.5 & 71.7 & 26.3 \\
\bottomrule
\end{tabular}
}
\caption{Performance of our method with varying $\alpha$ regularization parameter. Additional regularization has a negligible effect while avoiding potential trivial clustering.}
\label{tab:additional_regularization}
\end{table}

\subsubsection{Additional Regularization}
In addition to the primary and the auxiliary information objective terms, we propose another regularization term, the squared average node similarity, $ \alpha(\frac{1}{n^2}\sum_{ij} \langle X_iX_j \rangle)^2= \alpha\|\bar{X}\|_2^4$, where $\bar{X}$ is the mean of the node embedding vectors and $\alpha$ is a tunable parameter. This regularizer can be used to avoid the formation of a trivial clustering where all the nodes form a single cluster. Table \ref{tab:additional_regularization} shows that adding this regularizer have a negligible effect on our main objectives while avoiding potential trivial clustering.

\begin{figure}[ht]
    \centering
    \includegraphics[width=0.42\textwidth]{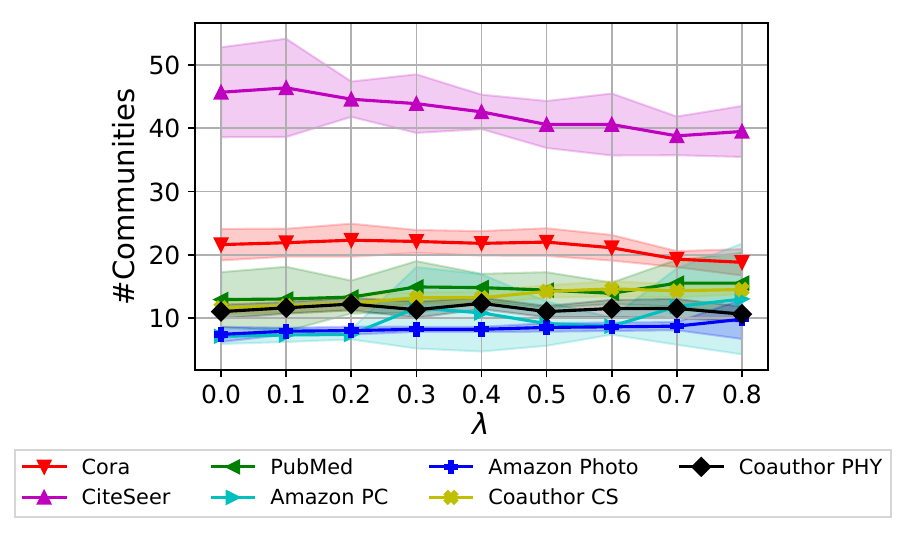}
    \caption{The number of communities varying $\lambda$ hyperparameter for different datasets. We observe that the number of communities found is similar across different $\lambda$ values except in some cases.}
    \label{fig:lambda_effect_on_number_of_communities}
\end{figure}

\subsection{Number of Communities}

Our method does not assume the number of communities in the data, which is one of its main strengths. The number of communities in real-world data is generally unknown, and it is nontrivial to estimate it. To share some insight, we provide the number of communities our method finds across different datasets and $\lambda$ values. Figure \ref{fig:lambda_effect_on_number_of_communities} demonstrates the number of communities varies for different datasets, and the results are consistent across different $\lambda$ values.

\section{Conclusion}
In this paper, we have studied the problem of graph clustering via maximizing modularity. The existing techniques often require the number of clusters beforehand or they fail to capitalize on the potential benefits of associated node attributes and availability of supplementary information. To address these, we have introduced {\model}, a novel neural framework that works well without a predefined number of clusters. Moreover, our framework uses graph neural networks (GNNs) to leverage the node attributes and additional information present within the graph. The computational complexity of {\model} scales linearly with the graph size.
Through extensive experimentation across seven real-world datasets of varying sizes and employing multiple distinct cluster quality evaluation metrics, we have showcased the superior performance of {\model}: it consistently has outperformed state-of-the-art approaches across most of the settings.

\clearpage
\section{Acknowledgments}
This material is based upon work partially supported by the National Science Foundation under grant no. 2229876 and in part by funds provided by the National Science Foundation, Department of Homeland Security, and IBM. 

\bibliography{aaai24}

\begin{thebibliography}{47}
\providecommand{\natexlab}[1]{#1}

\bibitem[{Bhatia and Rani(2018)}]{bhatia2018dfuzzy}
Bhatia, V.; and Rani, R. 2018.
\newblock Dfuzzy: a deep learning-based fuzzy clustering model for large graphs.
\newblock \emph{Knowledge and Information Systems}, 57: 159--181.

\bibitem[{Bianchi, Grattarola, and Alippi(2020)}]{bianchi2020spectral}
Bianchi, F.~M.; Grattarola, D.; and Alippi, C. 2020.
\newblock Spectral clustering with graph neural networks for graph pooling.
\newblock In \emph{International Conference on Machine Learning}, 874--883. PMLR.

\bibitem[{Blondel et~al.(2008)Blondel, Guillaume, Lambiotte, and Lefebvre}]{blondel2008fast}
Blondel, V.~D.; Guillaume, J.-L.; Lambiotte, R.; and Lefebvre, E. 2008.
\newblock Fast unfolding of communities in large networks.
\newblock \emph{Journal of statistical mechanics: theory and experiment}, 2008(10): P10008.

\bibitem[{Bo et~al.(2020)Bo, Wang, Shi, Zhu, Lu, and Cui}]{bo2020structural}
Bo, D.; Wang, X.; Shi, C.; Zhu, M.; Lu, E.; and Cui, P. 2020.
\newblock Structural deep clustering network.
\newblock In \emph{Proceedings of the web conference 2020}, 1400--1410.

\bibitem[{Brandes et~al.(2006)Brandes, Delling, Gaertler, G{\"o}rke, Hoefer, Nikoloski, and Wagner}]{brandes2006maximizing}
Brandes, U.; Delling, D.; Gaertler, M.; G{\"o}rke, R.; Hoefer, M.; Nikoloski, Z.; and Wagner, D. 2006.
\newblock Maximizing modularity is hard.
\newblock \emph{arXiv preprint physics/0608255}.

\bibitem[{Choong, Liu, and Murata(2018)}]{choong2018learning}
Choong, J.~J.; Liu, X.; and Murata, T. 2018.
\newblock Learning community structure with variational autoencoder.
\newblock In \emph{2018 IEEE international conference on data mining (ICDM)}, 69--78. IEEE.

\bibitem[{Felzenszwalb and Huttenlocher(2004)}]{felzenszwalb2004efficient}
Felzenszwalb, P.~F.; and Huttenlocher, D.~P. 2004.
\newblock Efficient graph-based image segmentation.
\newblock \emph{International journal of computer vision}, 59: 167--181.

\bibitem[{Fortunato and Hric(2016)}]{fortunato2016community}
Fortunato, S.; and Hric, D. 2016.
\newblock Community detection in networks: A user guide.
\newblock \emph{Physics reports}, 659: 1--44.

\bibitem[{Frey and Dueck(2007)}]{frey2007clustering}
Frey, B.~J.; and Dueck, D. 2007.
\newblock Clustering by passing messages between data points.
\newblock \emph{science}, 315(5814): 972--976.

\bibitem[{Hamilton, Ying, and Leskovec(2017)}]{hamilton2017inductive}
Hamilton, W.; Ying, Z.; and Leskovec, J. 2017.
\newblock Inductive representation learning on large graphs.
\newblock \emph{Advances in neural information processing systems}, 30.

\bibitem[{Hartigan and Wong(1979)}]{hartigan1979algorithm}
Hartigan, J.~A.; and Wong, M.~A. 1979.
\newblock Algorithm AS 136: A k-means clustering algorithm.
\newblock \emph{Journal of the royal statistical society. series c (applied statistics)}, 28(1): 100--108.

\bibitem[{Kipf and Welling(2016)}]{kipf2016semi}
Kipf, T.~N.; and Welling, M. 2016.
\newblock Semi-supervised classification with graph convolutional networks.
\newblock \emph{arXiv preprint arXiv:1609.02907}.

\bibitem[{Klambauer et~al.(2017)Klambauer, Unterthiner, Mayr, and Hochreiter}]{klambauer2017self}
Klambauer, G.; Unterthiner, T.; Mayr, A.; and Hochreiter, S. 2017.
\newblock Self-normalizing neural networks.
\newblock \emph{Advances in neural information processing systems}, 30.

\bibitem[{Kulatilleke, Portmann, and Chandra(2022)}]{kulatilleke2022scgc}
Kulatilleke, G.~K.; Portmann, M.; and Chandra, S.~S. 2022.
\newblock SCGC: Self-supervised contrastive graph clustering.
\newblock \emph{arXiv preprint arXiv:2204.12656}.

\bibitem[{Liu et~al.(2023)Liu, Yang, Zhou, Liu, Wang, Liang, Tu, and Li}]{liu2023simple}
Liu, Y.; Yang, X.; Zhou, S.; Liu, X.; Wang, S.; Liang, K.; Tu, W.; and Li, L. 2023.
\newblock Simple contrastive graph clustering.
\newblock \emph{IEEE Transactions on Neural Networks and Learning Systems}.

\bibitem[{Mandaglio, Amelio, and Tagarelli(2018)}]{mandaglio2018consensus}
Mandaglio, D.; Amelio, A.; and Tagarelli, A. 2018.
\newblock Consensus community detection in multilayer networks using parameter-free graph pruning.
\newblock In \emph{Advances in Knowledge Discovery and Data Mining: 22nd Pacific-Asia Conference, PAKDD 2018, Melbourne, VIC, Australia, June 3-6, 2018, Proceedings, Part III 22}, 193--205. Springer.

\bibitem[{Moradi, Ahmadian, and Akhlaghian(2015)}]{moradi2015effective}
Moradi, P.; Ahmadian, S.; and Akhlaghian, F. 2015.
\newblock An effective trust-based recommendation method using a novel graph clustering algorithm.
\newblock \emph{Physica A: Statistical mechanics and its applications}, 436: 462--481.

\bibitem[{Newman(2006{\natexlab{a}})}]{newman2006finding}
Newman, M.~E. 2006{\natexlab{a}}.
\newblock Finding community structure in networks using the eigenvectors of matrices.
\newblock \emph{Physical review E}, 74(3): 036104.

\bibitem[{Newman(2006{\natexlab{b}})}]{newman2006modularity}
Newman, M.~E. 2006{\natexlab{b}}.
\newblock Modularity and community structure in networks.
\newblock \emph{Proceedings of the national academy of sciences}, 103(23): 8577--8582.

\bibitem[{Newman and Girvan(2003)}]{newman2003mixing}
Newman, M.~E.; and Girvan, M. 2003.
\newblock Mixing patterns and community structure in networks.
\newblock In \emph{Statistical mechanics of complex networks}, 66--87. Springer.

\bibitem[{Park et~al.(2019)Park, Lee, Chang, Lee, and Choi}]{park2019symmetric}
Park, J.; Lee, M.; Chang, H.~J.; Lee, K.; and Choi, J.~Y. 2019.
\newblock Symmetric graph convolutional autoencoder for unsupervised graph representation learning.
\newblock In \emph{Proceedings of the IEEE/CVF international conference on computer vision}, 6519--6528.

\bibitem[{Peixoto(2014)}]{peixoto2014efficient}
Peixoto, T.~P. 2014.
\newblock Efficient Monte Carlo and greedy heuristic for the inference of stochastic block models.
\newblock \emph{Physical Review E}, 89(1): 012804.

\bibitem[{Perozzi, Al-Rfou, and Skiena(2014)}]{perozzi2014deepwalk}
Perozzi, B.; Al-Rfou, R.; and Skiena, S. 2014.
\newblock Deepwalk: Online learning of social representations.
\newblock In \emph{Proceedings of the 20th ACM SIGKDD international conference on Knowledge discovery and data mining}, 701--710.

\bibitem[{Scarselli et~al.(2008)Scarselli, Gori, Tsoi, Hagenbuchner, and Monfardini}]{scarselli2008graph}
Scarselli, F.; Gori, M.; Tsoi, A.~C.; Hagenbuchner, M.; and Monfardini, G. 2008.
\newblock The graph neural network model.
\newblock \emph{IEEE transactions on neural networks}, 20(1): 61--80.

\bibitem[{Sen et~al.(2008)Sen, Namata, Bilgic, Getoor, Galligher, and Eliassi-Rad}]{sen2008collective}
Sen, P.; Namata, G.; Bilgic, M.; Getoor, L.; Galligher, B.; and Eliassi-Rad, T. 2008.
\newblock Collective classification in network data.
\newblock \emph{AI magazine}, 29(3): 93--93.

\bibitem[{Shchur and G{\"u}nnemann(2019)}]{shchur2019overlapping}
Shchur, O.; and G{\"u}nnemann, S. 2019.
\newblock Overlapping community detection with graph neural networks.
\newblock \emph{arXiv preprint arXiv:1909.12201}.

\bibitem[{Shchur et~al.(2018)Shchur, Mumme, Bojchevski, and G{\"u}nnemann}]{shchur2018pitfalls}
Shchur, O.; Mumme, M.; Bojchevski, A.; and G{\"u}nnemann, S. 2018.
\newblock Pitfalls of graph neural network evaluation.
\newblock \emph{arXiv preprint arXiv:1811.05868}.

\bibitem[{Shi and Malik(2000)}]{shi2000normalized}
Shi, J.; and Malik, J. 2000.
\newblock Normalized cuts and image segmentation.
\newblock \emph{IEEE Transactions on pattern analysis and machine intelligence}, 22(8): 888--905.

\bibitem[{Sun et~al.(2021)Sun, Zheng, Zhang, and Xu}]{sun2021graph}
Sun, J.; Zheng, W.; Zhang, Q.; and Xu, Z. 2021.
\newblock Graph neural network encoding for community detection in attribute networks.
\newblock \emph{IEEE Transactions on Cybernetics}, 52(8): 7791--7804.

\bibitem[{Tsitsulin et~al.(2023)Tsitsulin, Palowitch, Perozzi, and M{\"u}ller}]{muller2023graph}
Tsitsulin, A.; Palowitch, J.; Perozzi, B.; and M{\"u}ller, E. 2023.
\newblock Graph clustering with graph neural networks.
\newblock \emph{Journal of Machine Learning Research}, 24(127): 1--21.

\bibitem[{Velickovic et~al.(2017)Velickovic, Cucurull, Casanova, Romero, Lio, Bengio et~al.}]{velickovic2017graph}
Velickovic, P.; Cucurull, G.; Casanova, A.; Romero, A.; Lio, P.; Bengio, Y.; et~al. 2017.
\newblock Graph attention networks.
\newblock \emph{stat}, 1050(20): 10--48550.

\bibitem[{Veli{\v{c}}kovi{\'c} et~al.(2018)Veli{\v{c}}kovi{\'c}, Fedus, Hamilton, Li{\`o}, Bengio, and Hjelm}]{velivckovic2018deep}
Veli{\v{c}}kovi{\'c}, P.; Fedus, W.; Hamilton, W.~L.; Li{\`o}, P.; Bengio, Y.; and Hjelm, R.~D. 2018.
\newblock Deep graph infomax.
\newblock \emph{arXiv preprint arXiv:1809.10341}.

\bibitem[{Wang et~al.(2019)Wang, Pan, Hu, Long, Jiang, and Zhang}]{wang2019attributed}
Wang, C.; Pan, S.; Hu, R.; Long, G.; Jiang, J.; and Zhang, C. 2019.
\newblock Attributed graph clustering: A deep attentional embedding approach.
\newblock \emph{arXiv preprint arXiv:1906.06532}.

\bibitem[{Wang et~al.(2017)Wang, Pan, Long, Zhu, and Jiang}]{wang2017mgae}
Wang, C.; Pan, S.; Long, G.; Zhu, X.; and Jiang, J. 2017.
\newblock Mgae: Marginalized graph autoencoder for graph clustering.
\newblock In \emph{Proceedings of the 2017 ACM on Conference on Information and Knowledge Management}, 889--898.

\bibitem[{Wang et~al.(2010)Wang, Li, Chen, and Pan}]{wang2010fast}
Wang, J.; Li, M.; Chen, J.; and Pan, Y. 2010.
\newblock A fast hierarchical clustering algorithm for functional modules discovery in protein interaction networks.
\newblock \emph{IEEE/ACM Transactions on Computational Biology and Bioinformatics}, 8(3): 607--620.

\bibitem[{Wu et~al.(2020)Wu, Zhang, Chen, Guo, and Wang}]{wu2020deep}
Wu, L.; Zhang, Q.; Chen, C.-H.; Guo, K.; and Wang, D. 2020.
\newblock Deep learning techniques for community detection in social networks.
\newblock \emph{IEEE Access}, 8: 96016--96026.

\bibitem[{Xia et~al.(2021)Xia, Wang, Gao, Zhang, and Gao}]{xia2021self}
Xia, W.; Wang, Q.; Gao, Q.; Zhang, X.; and Gao, X. 2021.
\newblock Self-supervised graph convolutional network for multi-view clustering.
\newblock \emph{IEEE Transactions on Multimedia}, 24: 3182--3192.

\bibitem[{Xu et~al.(2018)Xu, Hu, Leskovec, and Jegelka}]{xu2018powerful}
Xu, K.; Hu, W.; Leskovec, J.; and Jegelka, S. 2018.
\newblock How powerful are graph neural networks?
\newblock \emph{arXiv preprint arXiv:1810.00826}.

\bibitem[{Yang and Leskovec(2012)}]{yang2012defining}
Yang, J.; and Leskovec, J. 2012.
\newblock Defining and evaluating network communities based on ground-truth.
\newblock In \emph{Proceedings of the ACM SIGKDD Workshop on Mining Data Semantics}, 1--8.

\bibitem[{Yang et~al.(2016)Yang, Cao, He, Wang, Wang, and Zhang}]{yang2016modularity}
Yang, L.; Cao, X.; He, D.; Wang, C.; Wang, X.; and Zhang, W. 2016.
\newblock Modularity based community detection with deep learning.
\newblock In \emph{IJCAI}, volume~16, 2252--2258.

\bibitem[{Ying et~al.(2018{\natexlab{a}})Ying, He, Chen, Eksombatchai, Hamilton, and Leskovec}]{ying2018graph}
Ying, R.; He, R.; Chen, K.; Eksombatchai, P.; Hamilton, W.~L.; and Leskovec, J. 2018{\natexlab{a}}.
\newblock Graph convolutional neural networks for web-scale recommender systems.
\newblock In \emph{Proceedings of the 24th ACM SIGKDD international conference on knowledge discovery \& data mining}, 974--983.

\bibitem[{Ying et~al.(2018{\natexlab{b}})Ying, You, Morris, Ren, Hamilton, and Leskovec}]{ying2018hierarchical}
Ying, Z.; You, J.; Morris, C.; Ren, X.; Hamilton, W.; and Leskovec, J. 2018{\natexlab{b}}.
\newblock Hierarchical graph representation learning with differentiable pooling.
\newblock \emph{Advances in neural information processing systems}, 31.

\bibitem[{Yue et~al.(2022)Yue, Jun, Sihang, Siwei, Xifeng, Xihong, Ke, Wenxuan, Wang et~al.}]{yue2022survey}
Yue, L.; Jun, X.; Sihang, Z.; Siwei, W.; Xifeng, G.; Xihong, Y.; Ke, L.; Wenxuan, T.; Wang, L.~X.; et~al. 2022.
\newblock A survey of deep graph clustering: Taxonomy, challenge, and application.
\newblock \emph{arXiv preprint arXiv:2211.12875}.

\bibitem[{Zhang and Chen(2018)}]{zhang2018link}
Zhang, M.; and Chen, Y. 2018.
\newblock Link prediction based on graph neural networks.
\newblock \emph{Advances in neural information processing systems}, 31.

\bibitem[{Zhang, Ramakrishnan, and Livny(1996)}]{zhang1996birch}
Zhang, T.; Ramakrishnan, R.; and Livny, M. 1996.
\newblock BIRCH: an efficient data clustering method for very large databases.
\newblock \emph{ACM sigmod record}, 25(2): 103--114.

\bibitem[{Zhang et~al.(2019)Zhang, Liu, Li, and Wu}]{zhang2019attributed}
Zhang, X.; Liu, H.; Li, Q.; and Wu, X.-M. 2019.
\newblock Attributed graph clustering via adaptive graph convolution.
\newblock \emph{arXiv preprint arXiv:1906.01210}.

\bibitem[{Zhou et~al.(2020)Zhou, Cui, Hu, Zhang, Yang, Liu, Wang, Li, and Sun}]{zhou2020graph}
Zhou, J.; Cui, G.; Hu, S.; Zhang, Z.; Yang, C.; Liu, Z.; Wang, L.; Li, C.; and Sun, M. 2020.
\newblock Graph neural networks: A review of methods and applications.
\newblock \emph{AI open}, 1: 57--81.

\end{thebibliography}

\end{document}